\documentclass[letterpaper]{article} 
\usepackage{aaai2026}  
\usepackage{times}  
\usepackage{helvet}  
\usepackage{courier}  
\usepackage[hyphens]{url}  
\usepackage{graphicx} 
\usepackage{natbib}  
\usepackage{caption} 
\usepackage{algorithm}
\usepackage{algorithmic}

\usepackage{amsmath}
\usepackage{amssymb} 

\newcommand{\cmark}{\checkmark}

\usepackage{multirow}
\usepackage{booktabs}       
\usepackage{newfloat}
\usepackage{listings}
\DeclareCaptionStyle{ruled}{labelfont=normalfont,labelsep=colon,strut=off} 
\floatstyle{ruled}
\newfloat{listing}{tb}{lst}{}
\floatname{listing}{Listing}
\title{TG-Field: Geometry-Aware Radiative Gaussian Fields for Tomographic Reconstruction}
\author{
	Yuxiang Zhong\textsuperscript{\rm 1,2},
	Jun Wei\textsuperscript{\rm 1},
	Chaoqi Chen\textsuperscript{\rm 1},
	Senyou An\textsuperscript{\rm 2},
	Hui Huang\textsuperscript{\rm 1}\thanks{Corresponding author}
}

\affiliations{
    \textsuperscript{\rm 1}College of Computer Science and Software Engineering, Shenzhen University, Shenzhen, China\\

    \textsuperscript{\rm 2}State Key Laboratory of Intelligent Construction and Healthy Operation and Maintenance of Deep Underground Engineering, \\College of Civil and Transportation Engineering, Shenzhen University, Shenzhen, China\\
    \{yuxiangzhong0822, hhzhiyan\}@gmail.com, \{weijun, chencq, senyouan\}@szu.edu.cn
    
}

\usepackage{bibentry}

\begin{document}

\maketitle

\begin{abstract}
3D Gaussian Splatting (3DGS) has revolutionized 3D scene representation with superior efficiency and quality. While recent adaptations for computed tomography (CT) show promise, they struggle with severe artifacts under highly sparse-view projections and dynamic motions. To address these challenges, we propose Tomographic Geometry Field (TG-Field), a geometry-aware Gaussian deformation framework tailored for both static and dynamic CT reconstruction. A multi-resolution hash encoder is employed to capture local spatial priors, regularizing primitive parameters under ultra-sparse settings. We further extend the framework to dynamic reconstruction by introducing time-conditioned representations and a spatiotemporal attention block to adaptively aggregate features, thereby resolving spatiotemporal ambiguities and enforcing temporal coherence. In addition, a motion-flow network models fine-grained respiratory motion to track local anatomical deformations. Extensive experiments on synthetic and real-world datasets demonstrate that TG-Field consistently outperforms existing methods, achieving state-of-the-art reconstruction accuracy under highly sparse-view conditions.
\end{abstract}


\section{Introduction}

As a non-invasive imaging modality, Cone Beam Computed Tomography (CBCT) is widely used across medicine, biology, and industry to visualize internal structures~\cite{cormack1963representation, de2014industrial, kherlopian2008review}. 
To achieve high-quality 3D reconstructions, CBCT systems typically require hundreds of X-ray projections~\cite{feldkamp1984practical}.
However, such dense angular sampling raises concerns about radiation exposure, especially in repeated scans or high-dose protocols.
To mitigate radiation dose, sparse-view CBCT reconstruction has become a key research focus, aiming to reduce the number of projections while preserving reconstruction fidelity.

Traditional CT reconstruction methods typically rely on a discrete voxel-grid representation. Recently, 3D Gaussian Splatting (3DGS)~\cite{kerbl20233d} has emerged as a compelling alternative, modeling the volume with a set of anisotropic 3D Gaussians and supporting fast image formation through efficient splatting. Building on this idea, recent studies~\cite{cai2025radiative,gao2024ddgs,r2_gaussian} extend 3DGS to X-ray imaging, showing clear benefits over voxel-based baselines in accuracy and speed.

Nevertheless, existing methods still face two critical challenges in practical applications:
(1) \textbf{Robustness under extremely sparse views}.
While current methods perform well with moderately sparse projections, their quality sharply deteriorates when projections become extremely limited. Without explicit geometric regularization, per-splat optimization fails to maintain geometric consistency and structural coherence, leading to severe artifacts and loss of anatomical detail. This poses challenges for clinical adoption, where minimal-view imaging is crucial for dose reduction.
(2) \textbf{Struggle with dynamic CT reconstruction}.
Most existing methods are designed for static cases and struggle with respiratory motion. Modeling temporal coherence and capturing non-rigid deformations remain challenging, limiting effectiveness in real-world 4D CT.

To address these issues, we propose TG-Field, a geometry-aware Gaussian deformation framework designed for high-fidelity static and dynamic CT reconstruction. 
Specifically, we employ a multi-resolution hash encoder coupled with a two-stage iterative refinement strategy to capture local geometric context and ensure spatial continuity. 
To further tackle the ill-posedness of sparse-view projections, we introduce a semantic-consistency regularization module powered by pretrained visual foundation models, enforcing cross-view coherence.
For dynamic scenes, TG-Field embeds time-conditioned representations within each Gaussian. Since naïve joint spatiotemporal hashing often induces collisions and temporal drift, we incorporate a spatiotemporal attention block and a motion-flow network.


Our main contributions are summarized as follows: 
\begin{itemize}
	\item A new Gaussian deformation framework that achieves high-quality static and dynamic reconstruction under ultra-sparse views.
	\item Several technical designs, including a hash-grid encoder, a spatiotemporal attention module, and a motion-flow network to model dynamic local deformations, plus cross-view semantic-consistency regularization.
	\item Experiments across diverse CT scenarios demonstrate the superiority of TG-Field over existing methods, highlighting robustness and potential for clinical workflows.
	
\end{itemize}

\section{Related Work}

\subsection{Cone-Beam CT Reconstruction}

CBCT reconstruction methods are commonly categorized as analytical and iterative. Analytical approaches~\cite{feldkamp1984practical, radon1986determination} provide fast, closed-form inversion but degrade markedly under sparse-view sampling, whereas iterative approaches~\cite{andersen1984simultaneous, sidky2008image} improve robustness by solving data-fidelity–regularization objectives at the cost of higher computation and increased sensitivity to hyperparameter settings and prior selection. Recently, deep learning–based methods have demonstrated substantial improvements across CBCT tasks, including projection completion~\cite{anirudh2018lose, ghani2018deep}, volume enhancement~\cite{chung2023solving, lee2023improving, liu2023dolce, liu2020tomogan}, and direct volume reconstruction from sparse-view or limited-angle data~\cite{adler2018learned, jin2017deep, lin2023learning, lin2024c, ying2019x2ct}. Nevertheless, these supervised approaches often struggle to generalize to out-of-distribution scenarios due to their reliance on paired training data.

To mitigate such limitations, recent advances leverage self-supervised, NeRF-inspired implicit representations~\cite{mildenhall2021nerf}, enabling CT reconstruction without paired supervision. Representative methods include IntraTomo~\cite{zang2021intratomo}, NAF~\cite{zha2022naf}, SAX-NeRF~\cite{cai2024structure}, NEAT~\cite{ruckert2022neat}, and NeRP~\cite{shen2022nerp}. These methods optimize continuous implicit fields for reconstruction. However, most NeRF-based approaches focus on static reconstruction and require dense ray sampling, incurring significant computational cost. Recently, STNF4D~\cite{zhou2025spatiotemporal} adopts a structured grid representation to model dynamic CT scans, yet it still converges slowly and yields suboptimal reconstruction quality.

\begin{figure}[t]
	\centering
	\includegraphics[width=1\linewidth]{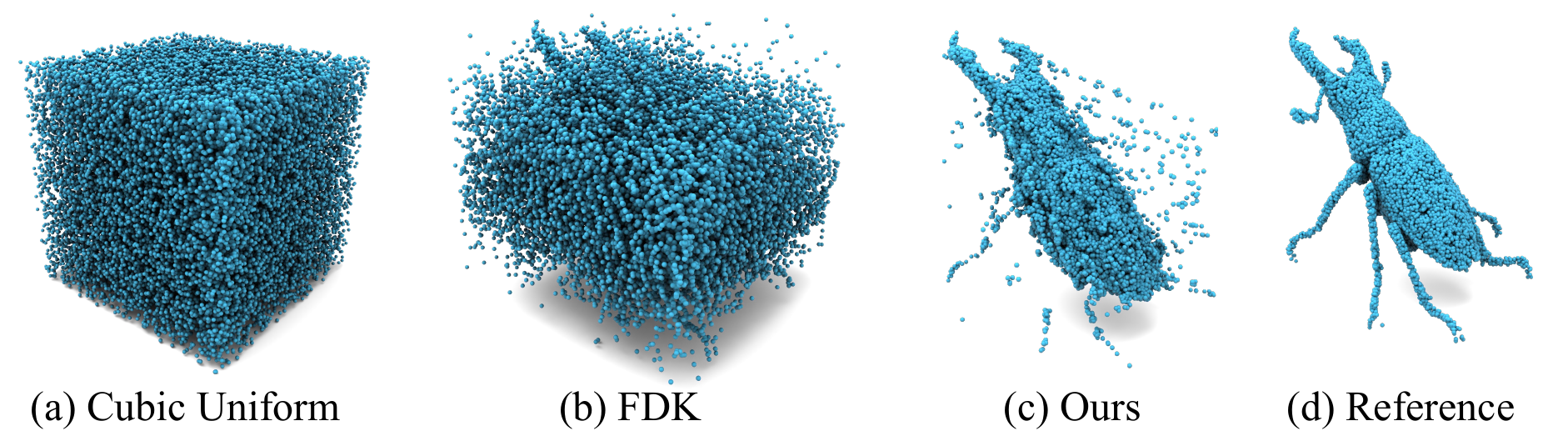}
	\caption{
		Visual comparison of point clouds obtained by different initialization methods:
		(a) cubic uniform sampling ~\cite{cai2025radiative},
		(b) FDK-based initialization ~\cite{r2_gaussian},
		(c) our iteration-based initialization, and
		(d) reference.
		Our method clearly captures more accurate and detailed geometric structures.
	}
	\label{fig:init_comparison}
\end{figure}

\subsection{3D Gaussian Splatting in Medical Imaging}
3D Gaussian Splatting (3DGS)~\cite{kerbl20233d} enables real-time rendering and rapid convergence, and has recently been adapted to medical imaging~\cite{cai2025radiative, gao2024ddgs, r2_gaussian}. For synthesizing X-ray projections, X-Gaussian~\cite{cai2025radiative} introduces grayscale modeling, while DDGS~\cite{gao2024ddgs} incorporates radiosity decomposition. In contrast, $\text{R}^2$-Gaussian~\cite{r2_gaussian} directly targets CT volume reconstruction via differentiable voxelization. Building on these static formulations, Spatiotemporal Gaussian~\cite{fu2025spatiotemporal} extends $\text{R}^2$-Gaussian with a deformation network and parameterizes the representation with HexPlane to construct a spatiotemporal Gaussian field for dynamic CBCT; $\text{X}^2$-Gaussian~\cite{yu2025x} also adopts a HexPlane representation and integrates dynamic radiative GS with self-supervised respiratory motion learning. While effective under moderate sampling, these dynamic models degrade as views become sparser. In contrast, TG-Field explicitly targets robustness under ultra-sparse views and provides a geometry-aware framework for both static and dynamic CT.

\section{Preliminaries}

Radiative Gaussian Splatting~\cite{r2_gaussian} models the 3D CT density field as a finite sum of anisotropic Gaussian kernels \(\mathbb{G}=\{G_i\}_{i=1}^{N}\).
Each kernel \(G_i\) is parameterized by density \(\rho_i\), center \(\boldsymbol{\mu}_i\in\mathbb{R}^3\), and covariance \(\Sigma_i\in\mathbb{R}^{3\times3}\):
\begin{equation*}
  G_i(\mathbf{x}\mid\rho_i,\boldsymbol{\mu}_i,\Sigma_i)
  = \rho_i \exp\!\left(-\tfrac{1}{2}(\mathbf{x}-\boldsymbol{\mu}_i)^{\top}\Sigma_i^{-1}(\mathbf{x}-\boldsymbol{\mu}_i)\right).
\end{equation*}
The density field is the superposition
\begin{equation*}
  \sigma(\mathbf{x})=\sum_{i=1}^{N} G_i(\mathbf{x}\mid\rho_i,\boldsymbol{\mu}_i,\Sigma_i).
\end{equation*}

Let the X-ray be \(\mathbf{r}(t)=\mathbf{o}+t\mathbf{d}\) for \(t\in[t_{\mathrm{in}},t_{\mathrm{out}}]\).
By the Beer–Lambert law~\cite{swinehart1962beer}, the detected intensity is
\begin{equation*}
  I'(\mathbf{r}) = I_0 \exp\!\left(-\!\int_{t_{\mathrm{in}}}^{t_{\mathrm{out}}}\sigma(\mathbf{r}(t))\,dt\right),
\end{equation*}
and the log-transformed measurement (line integral) is
\begin{equation*}
  I(\mathbf{r}) = \log I_0 - \log I'(\mathbf{r})
  = \int_{t_{\mathrm{in}}}^{t_{\mathrm{out}}}\sigma(\mathbf{r}(t))\,dt.
\end{equation*}
Thus each projection pixel \(I(\mathbf{r})\) records the integral of \(\sigma\) along its ray.

\begin{figure*}[!ht]
	\centering
	\includegraphics[width=\textwidth]{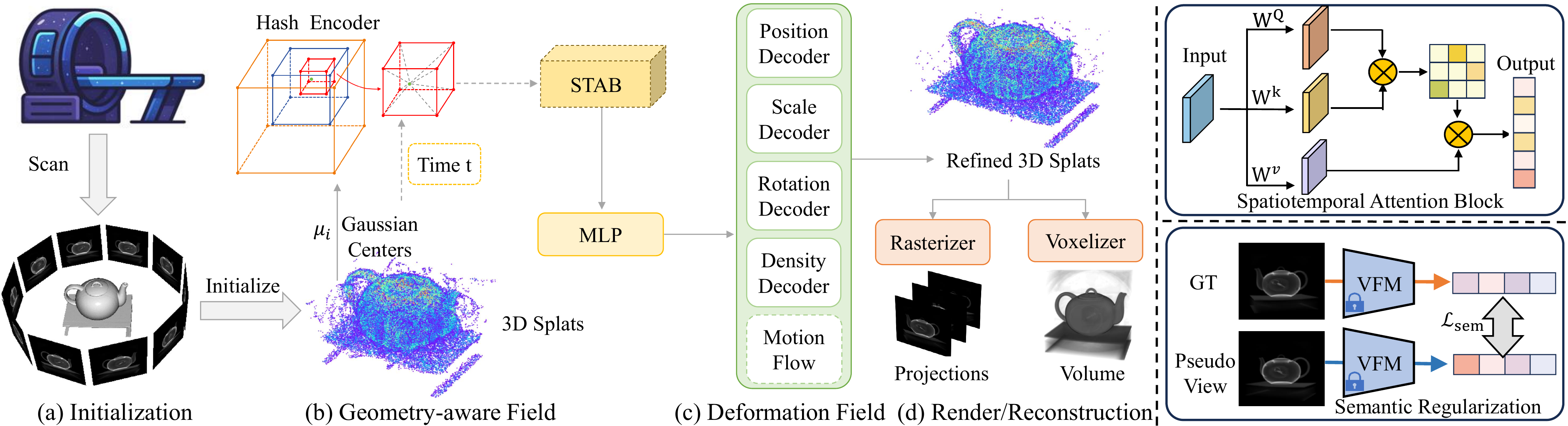}
	\caption{Our TG-Field framework first initializes a point cloud via a two-stage iterative refinement. The refined coordinates are encoded using a hash encoder to capture spatial geometric features, which are then decoded into optimized Gaussian splats by multiple MLP decoders. Additionally, we employ semantic consistency regularization with pretrained visual foundation models to maintain multi-view semantic
		consistency. Finally, these Gaussian splats are rendered into arbitrary-view X-ray projections and voxelized into a CT volume. For dynamic scenarios, we incorporate a spatiotemporal attention module and a Motion-Flow
		module to effectively capture temporal variations and fine-grained local movements.} 
	\label{fig:network}
\end{figure*}

\section{Method}
\label{method}
We begin with a iterative strategy that produces a structurally faithful initial point cloud. We then introduce a geometry-aware deformation framework that regularizes Gaussian primitives via a hash-grid encoder. Next, we extend the framework to dynamic CT by injecting time-conditioned representations together with a spatiotemporal attention module and a motion-flow network to capture respiratory motion. Finally, we impose a  semantic-consistency regularization to further enhance reconstruction quality.

\subsection{Iteration-based Initialization}
\label{IBI}
High-quality point cloud initialization is crucial for ensuring fast convergence and high-fidelity reconstruction in 3DGS-based methods~\cite{Wu_2024_CVPR}. However, under sparse-view CT settings, existing point cloud initialization strategies such as cubic grid sampling~\cite{cai2025radiative} and FDK-based methods~\cite{r2_gaussian} struggle to preserve the geometric structures of the target, resulting in degraded reconstructions.

To address this, we propose an iteration-based initialization consisting of two iterative stages. First, we use the algebraic reconstruction method Conjugate Gradient Least Squares~\cite{hestenes1952methods} to iteratively obtain a coarse volumetric reconstruction from sparse projections. Subsequently, we refine this coarse reconstruction using the regularization-based Adaptive Steepest Descent-Projection Onto Convex Sets~\cite{sidky2008image}, enforcing total variation (TV) constraints to reduce noise and preserve structural edges. As shown in Figure~\ref{fig:init_comparison}-c, this iterative approach significantly improves initialization quality. The refined volume is then converted into Gaussian primitives characterized by position, covariance, and attenuation derived from voxel intensity, further optimized via our geometry-aware neural field.

\begin{table*}[t]
	\centering
        \scriptsize 
		\begin{tabular}{c|c c|c c|c c|c c|c c|c c}
			\specialrule{1.5pt}{0pt}{0pt}
			\multirow{2}{*}{Method}
			& \multicolumn{2}{c|}{Synthetic (5 Views)} 
			& \multicolumn{2}{c|}{Synthetic (10 Views)} 
			& \multicolumn{2}{c|}{Synthetic (20 Views)} 
			& \multicolumn{2}{c|}{Real (5 Views)} 
			& \multicolumn{2}{c|}{Real (10 Views)} 
			& \multicolumn{2}{c}{Real (20 Views)} \\
			\cmidrule(lr){2-3} \cmidrule(lr){4-5} \cmidrule(lr){6-7} \cmidrule(lr){8-9} \cmidrule(lr){10-11} \cmidrule(lr){12-13}
			& PSNR $\uparrow$ & SSIM $\uparrow$ & PSNR $\uparrow$ & SSIM $\uparrow$ & PSNR $\uparrow$ & SSIM $\uparrow$ & PSNR $\uparrow$ & SSIM $\uparrow$ & PSNR $\uparrow$ & SSIM $\uparrow$ & PSNR $\uparrow$ & SSIM $\uparrow$ \\
			\midrule
			FDK 
			& 11.83 & 0.112 & 15.21 & 0.186 & 18.48 & 0.293 & 13.20 & 0.109 & 17.57 & 0.225 & 21.03 & 0.356 \\
			SART 
			& 22.10 & 0.683 & 24.32 & 0.768 & 27.24 & 0.845 & 26.22 & 0.771 & 28.72 & 0.846 & 30.94 & 0.899 \\
			IntraTomo 
			& 23.79 & 0.733 & 26.47 & 0.781 & 28.54 & 0.825 & 26.59 & 0.748 & 29.67 & 0.802 & 31.65 & 0.835 \\
			NAF
			& 23.83 & 0.711 & 27.41 & 0.784 & 30.59 & 0.849 & 26.76 & 0.769 & 30.94 & 0.814 & 33.71 & 0.853 \\
			SAX-NeRF
			& \underline{24.05} & \underline{0.740} & 27.55 & 0.801 & 31.93 & 0.875 & 26.96 & 0.784 & 32.26 & 0.835 & 35.07 & 0.863 \\
			X-Gaussian 
			& 23.62 & 0.699 & 27.59 & 0.804 & 31.79 & 0.868 & 26.71 & 0.766 & 31.04 & 0.819 & 34.91 & 0.852 \\
			$\text{R}^{2}$-Gaussian 
			& 23.81 & 0.735 & \underline{28.15} & \underline{0.833} & \underline{32.25} & \underline{0.923} & \underline{27.00} & \underline{0.789} & \underline{32.73} & \underline{0.859} & \underline{36.59} & \underline{0.901} \\
			Ours
			& \textbf{24.54} & \textbf{0.779} & \textbf{28.95} & \textbf{0.849} & \textbf{32.92} & \textbf{0.936} & \textbf{27.65} & \textbf{0.794} & \textbf{33.59} & \textbf{0.872} & \textbf{37.20} & \textbf{0.911} \\
			\specialrule{1.5pt}{0pt}{0pt}
		\end{tabular}%
        \caption{Quantitative comparisons of sparse-view static CT reconstruction on synthetic and real-world datasets. We \textbf{bold} the best results and \underline{underline} the second-best.}
	\label{tab:ct_recon}
	
\end{table*}

\subsection{Geometry-aware Splat Field}
\label{GAF}

We introduce a geometry-aware deformation field $\mathcal{D}$ that models spatial correlations among Gaussian primitives. By enforcing coherent deformation among the primitives, this framework compensates for missing geometric information under ultra-sparse projections and reduces artifacts.

Specifically, the deformation field predicts deformation parameters for each Gaussian primitive $G_i$, adjusting their positions and shapes to preserve geometric consistency. We write $G_i\equiv(\boldsymbol{\mu}_i,R_i,S_i,\rho_i)$. The deformed Gaussians $G'_i$ are computed as
\begin{equation}
  G'_i = (\boldsymbol{\mu}_i + \Delta\boldsymbol{\mu}_i,\; R_i + \Delta R_i,\; S_i + \Delta S_i,\; \rho_i + \Delta \rho_i),
\end{equation}
where $\Delta \boldsymbol{\mu}_i$, $\Delta R_i$, $\Delta S_i$, and $\Delta \rho_i$ denote offsets for position, rotation, scale, and density, respectively. Formally, the deformation field is defined as
\begin{equation}
  \mathcal{D} = \mathcal{F}\circ\mathcal{E}_{\text{Hash}},
\end{equation}
where $\mathcal{E}_{\text{Hash}}$ is a hash-grid encoder capturing spatial correlations, and $\mathcal{F}$ is a lightweight decoder.

\noindent \textbf{Spatially Decomposed Encoder.}
To explicitly capture spatial correlations under ultra-sparse views, we adopt a compact spatial decomposition with multiple low-dimensional feature grids inspired by HexPlane~\cite{cao2023hexplane} and 4DGS~\cite{wu20244d}. Concretely, the hash encoder is implemented as a set of trainable multi-resolution grids $\{G_s\}_{s\in S}$ that are queried at primitive coordinates. Given a primitive at coordinate $\boldsymbol{\mu}_i\in\mathbb{R}^3$, we obtain per-scale features by differentiable interpolation
\begin{equation}
  f_s(\boldsymbol{\mu}_i)=F(G_s,\boldsymbol{\mu}_i),\quad f_s(\boldsymbol{\mu}_i)\in\mathbb{R}^{C},
\end{equation}
and define the hash-encoded embedding as the multi-scale concatenation
\begin{equation}
  h_{\phi}(\boldsymbol{\mu}_i)=\operatorname{concat}_{s\in S}[f_s(\boldsymbol{\mu}_i)]\in\mathbb{R}^{|S|\cdot C}.
\end{equation}
This encoder provides multi-scale spatial context for each primitive and serves as the input to the deformation decoder.

\noindent\textbf{Multi-head Gaussian Deformation Decoder.}
Given $h_{\phi}(\boldsymbol{\mu}_i)$, a lightweight multi-head decoder outputs $\Delta \boldsymbol{\mu}_i=\mathcal{F}_{\mu}(h_{\phi}(\boldsymbol{\mu}_i)),\;
 \Delta R_i=\mathcal{F}_{R}(h_{\phi}(\boldsymbol{\mu}_i)),\;
 \Delta S_i=\mathcal{F}_{S}(h_{\phi}(\boldsymbol{\mu}_i)),\;
 \Delta \rho_i=\mathcal{F}_{\rho}(h_{\phi}(\boldsymbol{\mu}_i)).$

\noindent \textbf{4D CT Reconstruction.}
Our deformation-based framework naturally extends to dynamic 4D CT, addressing artifacts caused by respiratory motion and anatomical variations. We jointly encode spatial coordinates $\boldsymbol{\mu}_i$ and time $t$ to obtain per-scale features
\begin{equation}
  f_s(\boldsymbol{\mu}_i,t)=F\big(G_s,(\boldsymbol{\mu}_i,t)\big),\quad f_s(\boldsymbol{\mu}_i,t)\in\mathbb{R}^{C},
\end{equation}
and form a unified spatiotemporal embedding by concatenation
\begin{equation}
  h_{\phi}(\boldsymbol{\mu}_i,t)=\operatorname{concat}_{s\in S}[f_s(\boldsymbol{\mu}_i,t)]\in\mathbb{R}^{|S|\cdot C}.
\end{equation}

However, jointly hashing spatial and temporal coordinates can induce hash collisions, especially when identical (or near-identical) spatial locations reoccur across time, causing ambiguous embeddings and unreliable deformations.

To mitigate this, we recalibrate features across time with a spatiotemporal attention block (STAB). For each primitive $i$, we stack a short temporal window of embeddings
\begin{equation}
\mathbf{H}_i=\big[h_{\phi}(\boldsymbol{\mu}_i,t_1),\ldots,h_{\phi}(\boldsymbol{\mu}_i,t_T)\big]^{\top}\in\mathbb{R}^{T\times|S|\cdot C}.
\end{equation}
We compute queries, keys, and values as
\begin{equation}
Q=\mathbf{H}_i W_Q,\quad K=\mathbf{H}_i W_K,\quad V=\mathbf{H}_i W_V,
\end{equation}
where $W_Q,W_K,W_V\in\mathbb{R}^{C\times C}$, and we apply scaled dot-product attention
\begin{equation}
\mathrm{Attn}(\mathbf{H}_i)=\mathrm{softmax}\!\left(\frac{QK^{\top}}{\sqrt{C}}\right)V.
\end{equation}
The recalibrated embeddings are $\tilde{\mathbf{H}}_i=\mathbf{H}_i+\mathrm{Attn}(\mathbf{H}_i)$, from which we take the row corresponding to time $t$ as $\tilde{h}_{\phi}(\boldsymbol{\mu}_i,t)$. By aggregating temporal context, STAB disambiguates colliding bins at repeated spatial locations and yields more stable, accurate dynamic deformations. We then decode the per-primitive embedding at time $t$
 to obtain time-varying offsets:
\begin{equation}
\begin{aligned}
\Delta \boldsymbol{\mu}_i(t) &= \mathcal{F}_{\mu}\!\big(\tilde{h}_{\phi}(\boldsymbol{\mu}_i,t)\big), &
\Delta R_i(t) &= \mathcal{F}_{R}\!\big(\tilde{h}_{\phi}(\boldsymbol{\mu}_i,t)\big), \\
\Delta S_i(t) &= \mathcal{F}_{S}\!\big(\tilde{h}_{\phi}(\boldsymbol{\mu}_i,t)\big), &
\Delta \rho_i(t) &= \mathcal{F}_{\rho}\!\big(\tilde{h}_{\phi}(\boldsymbol{\mu}_i,t)\big).
\end{aligned}
\end{equation}

To address non-rigid tissue motion in 4D CT, we add a motion-flow module that predicts fine-grained displacement fields to further refine positions. We adopt a ResFields MLP~\cite{mihajlovic2024ResFields} to capture subtle local deformations that the initial deformation may miss, yielding more accurate Gaussian centers. Formally,
\begin{equation}
\hat{\boldsymbol{\mu}}_i(t)
= \boldsymbol{\mu}_i + \Delta\boldsymbol{\mu}_i(t)
  + \mathrm{Flow}\big(\boldsymbol{\mu}_i + \Delta\boldsymbol{\mu}_i(t),\, t\big),
\end{equation}
where $\mathrm{Flow}(\cdot,t)\in\mathbb{R}^3$ outputs a per-point displacement at time $t$.

\subsection{Optimization}
\label{optimization}
\textbf{Semantic Regularization.}
Prior work shows that coupling neural scene representations with visual foundation models (VFMs) (e.g., CLIP~\cite{radford2021learning} or DINOv2~\cite{oquab2023dinov2}) improves 3D semantic consistency and robustness~\cite{kerr2023lerf, qin2024langsplat}.
Under sparse views, purely photometric supervision is under-constrained; semantically unrelated textures can explain the projections equally well.
We therefore enforce cross-view semantic consistency using frozen VFM features on geometrically corresponding image regions.
Specifically, we render novel side views and extract VFM embeddings from their reprojected crops in a side view and a training view.
Let $P_i'$ and $P_i$ be a pair of corresponding crops for the $i$-th 3D location; we minimize
\begin{equation}
  \mathcal{L}_{\text{sem}}=\frac{1}{N}\sum_{i=1}^{N}\left\lVert f_{\text{vfm}}(P_i')-f_{\text{vfm}}(P_i)\right\rVert_2^{2},
\end{equation}
where $f_{\text{vfm}}$ is a pretrained VFM encoder and $\|\cdot\|_2$ denotes the $L_2$ norm.

\noindent \textbf{Loss Function.} We optimize our framework with a compound objective. We use an  L1 loss and D-SSIM to supervise the rendered X-ray projections. Following~\cite{r2_gaussian}, we additionally employ a 3D TV regularizer $\mathcal{L}_{\text{TV}}$ as a homogeneity prior for tomography. The overall loss is
\begin{equation}
  \mathcal{L}_{\text{total}} 
  = \mathcal{L}_1
  + \lambda_{\text{SSIM}}\,\mathcal{L}_{\text{SSIM}}
  + \lambda_{\text{TV}}\,\mathcal{L}_{\text{TV}}
  + \lambda_{\text{sem}}\,\mathcal{L}_{\text{sem}},
\end{equation}
with weighting factors $\lambda_{\text{SSIM}}$, $\lambda_{\text{TV}}$, and $\lambda_{\text{sem}}$ controlling the contribution of each component.

\begin{figure}[t]
    \centering
    \includegraphics[width=1\linewidth]{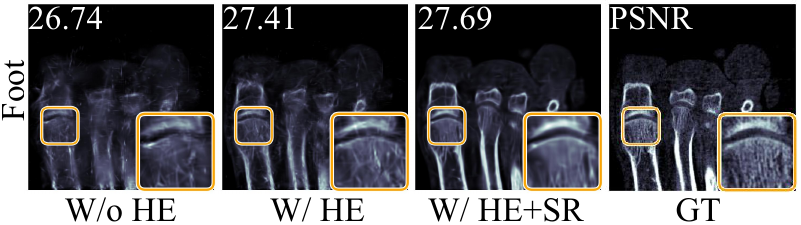}
    \caption{Ablation of the hash encoder (HE) and semantic regularization (SR) on static CT reconstruction.}
    \label{Aba}
\end{figure}

\noindent \textbf{Progressive Training Procedure.} During training, we first train an X-ray Radiative Gaussian splatting model~\cite{r2_gaussian} for 5000 iterations. This warm-up phase ensures the model effectively captures underlying anatomical structures from limited projection data. After warm-up, we incorporate our geometry-aware deformation field for refinement.

\begin{figure*}[t]
	\centering
	\includegraphics[width=1\linewidth]{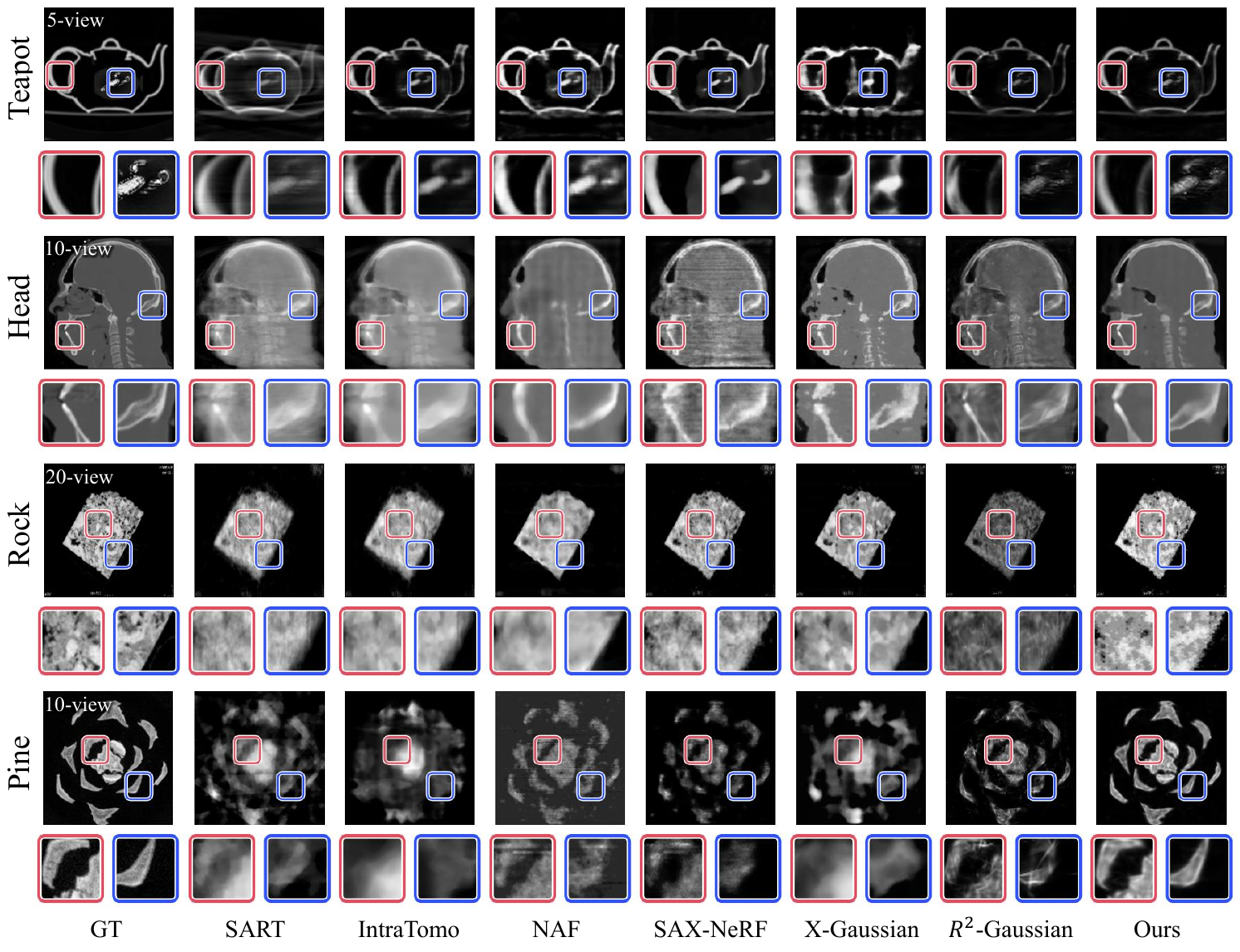}
	\caption{Qualitative comparisons of static CT reconstruction results across different view numbers (5, 10, and 20 views). Compared to SAX-NeRF and $\text{R}^2$-Gaussian, our method consistently provides sharper structural details and fewer artifacts, especially under extremely sparse-view conditions.}
	\label{view}
\end{figure*}

\begin{table*}[t]
    \centering
    \begin{tabular}{c*{4}{cc}}
        \toprule
        \multirow{2}{*}{\textbf{Method}} 
        & \multicolumn{2}{c}{XCAT}
        & \multicolumn{2}{c}{TCIA}
        & \multicolumn{2}{c}{SPARE}
        & \multicolumn{2}{c}{Average} \\
        \cmidrule(lr){2-3} \cmidrule(lr){4-5} \cmidrule(lr){6-7} \cmidrule(lr){8-9}
        & PSNR $\uparrow$ & SSIM $\uparrow$
        & PSNR $\uparrow$ & SSIM $\uparrow$
        & PSNR $\uparrow$ & SSIM $\uparrow$
        & PSNR $\uparrow$ & SSIM $\uparrow$ \\
        \midrule
        Hex-plane 
        & 21.79 & 0.866 
        & 23.91 & 0.835 
        & 26.43 & 0.856 
        & 24.04 & 0.852 \\
        
        K-plane 
        & 20.57 & 0.847 
        & 24.59 & 0.855 
        & 26.59 & 0.876 
        & 23.92 & 0.859 \\
        
        STNF4D
        & 25.73 & 0.928 
        & 29.37 & 0.919 
        & 28.75 & 0.887 
        & 27.95 & 0.911 \\
        
        4DGS 
        & \underline{33.95} & \underline{0.955} 
        & \underline{34.44} & \underline{0.948} 
        & \underline{30.01} & \underline{0.898} 
        & \underline{32.80} & \underline{0.933} \\
        
        Ours
        & \textbf{35.51} & \textbf{0.969} 
        & \textbf{35.41} & \textbf{0.955} 
        & \textbf{30.41} & \textbf{0.905} 
        & \textbf{33.78} & \textbf{0.943} \\
        
        \bottomrule
    \end{tabular}%
    \caption{Quantitative comparisons of our TG-Field with other methods on the 4D CT datasets. 
    We \textbf{bold} the best results and \underline{underline} the second-best.}
    \label{tab:dir_comparison}
\end{table*}

\begin{figure}[t]
    \centering
    \includegraphics[width=1\linewidth]{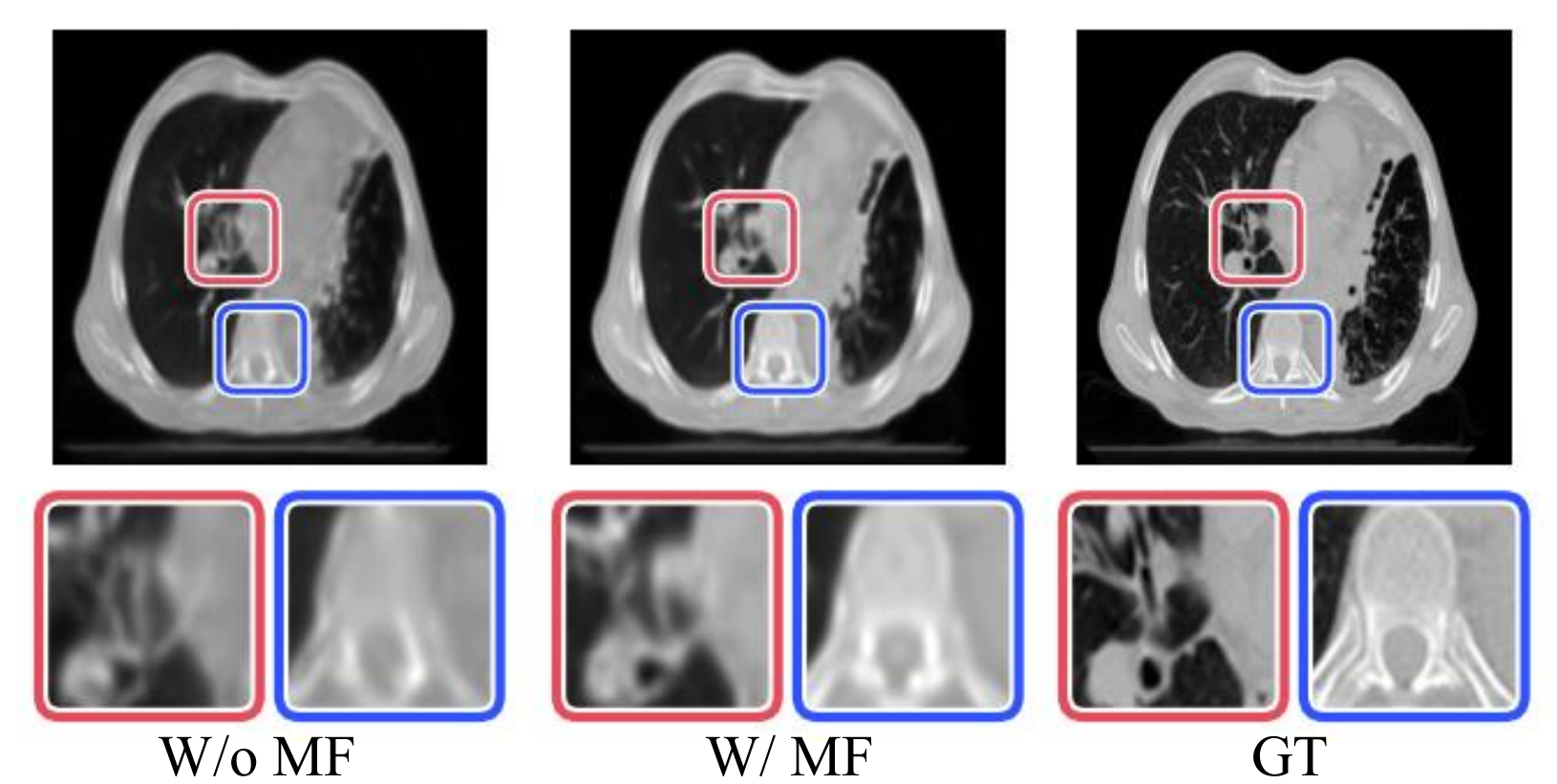}
    \caption{Visual ablation of the motion-flow module (MF) for 4D CT reconstruction.}
    \label{MF}
\end{figure}

\section{Experiments}
\label{experiment}
\subsection{Experiment Settings}\label{setting}

\textbf{Dataset.} We conduct experiments on both synthetic and real-world datasets, covering static and dynamic CT scans.
For static CT reconstruction, we evaluate our method on both synthetic~\cite{cai2024structure} and real data~\cite{r2_gaussian} under three sparse-view settings (5, 10, and 20 views).
For dynamic CT reconstruction, we use the 4D extended cardiac torso (XCAT) phantom~\cite{segars2008realistic}, real-patient 4D CT images from the 4D-Lung collection of TCIA~\cite{hugo2017longitudinal} (for simulated sparse-view projections), and clinically acquired 4D CBCT data from the SPARE challenge~\cite{shieh2019spare}.
For the XCAT and 4D-Lung datasets, each 4D CT sequence consists of 10 respiratory phases, and we simulate a total of 100 projections, evenly distributed across the phases, which is significantly fewer than the hundreds or thousands typically used in clinical protocols.
These X-ray projections are synthesized using the TIGRE tomography toolbox~\cite{biguri2016tigre} with a full $360^\circ$ circular trajectory and added noise to mimic realistic acquisition.

\begin{figure*}[t]
	\centering
	\includegraphics[width=1\linewidth]{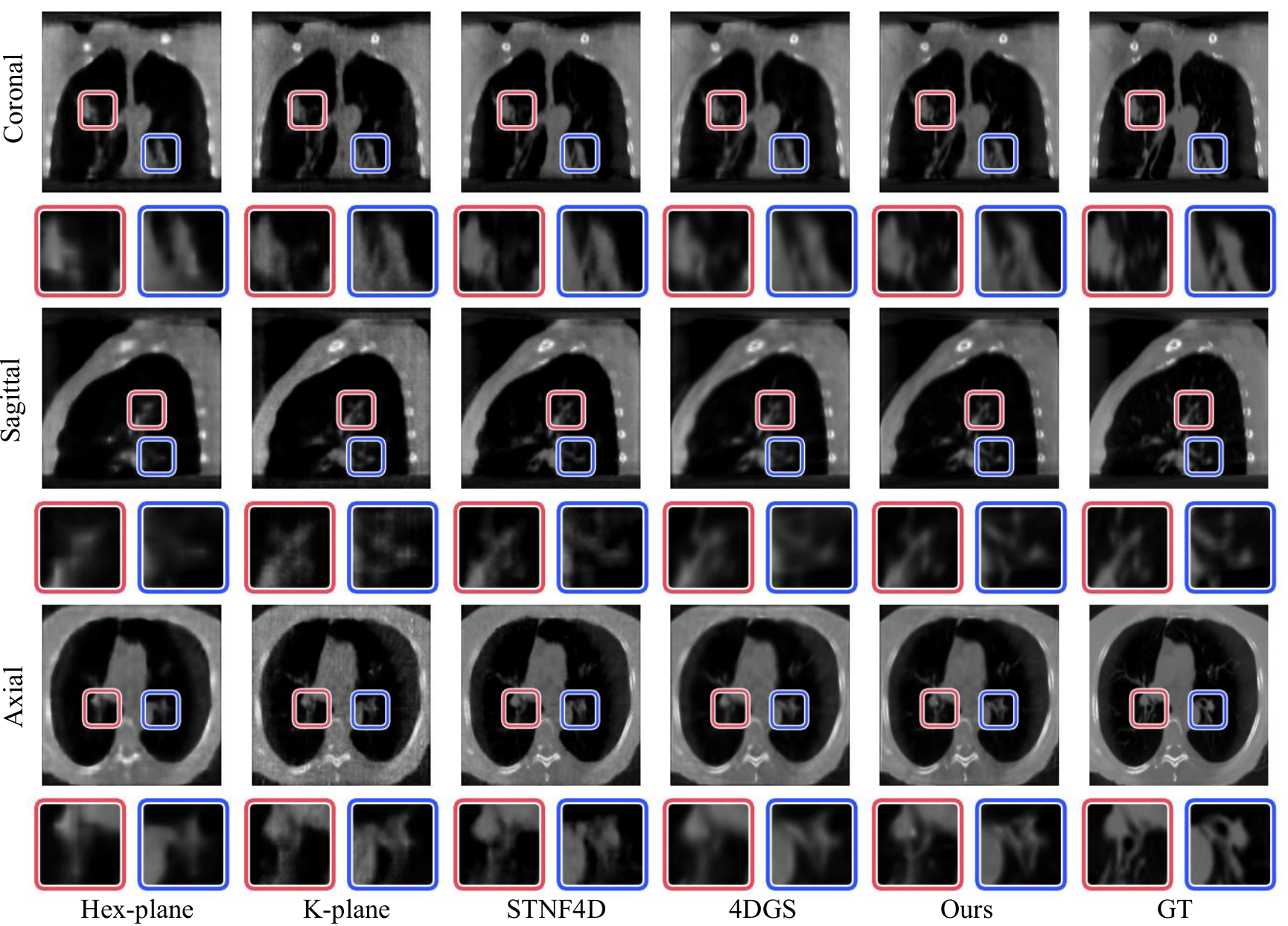}
	\caption{Visual comparisons of dynamic CT reconstruction results on the SPARE dataset. Our TG-Field shows superior performance in modeling anatomical details compared to existing approaches.}
	\label{4DCT}
\end{figure*}

\noindent \textbf{Implementation Details.}
TG-Field is implemented in PyTorch~\cite{paszke2019pytorch} with CUDA~\cite{sanders2010cuda}.
Training is performed on an NVIDIA L40 GPU using the Adam optimizer~\cite{kinga2015method} for 30k iterations.
The initial learning rates for the Gaussian parameters are set as follows: position at $2\times10^{-4}$, density at $1\times10^{-2}$, scale at $5\times10^{-3}$, and rotation at $1\times10^{-3}$, all exponentially decaying to $10\%$ of their initial values.
The hash encoder and decoder use initial learning rates of $2\times10^{-3}$ and $2\times10^{-4}$, respectively, also decaying to $10\%$ of their original values.
Loss weights are set to $\lambda_{\text{SSIM}}=0.25$, $\lambda_{\text{TV}}=0.05$, and $\lambda_{\text{sem}}=0.1$.
We adopt a pretrained DINO-ViT~\cite{zhang2022dino} model as the VFM for computing the semantic consistency loss.
During evaluation, we report PSNR and SSIM~\cite{wang2004image} to measure volumetric reconstruction quality.

\noindent \textbf{Baselines.}
For static CT reconstruction, we compare our method with several state-of-the-art baselines.
Traditional methods include FDK~\cite{feldkamp1984practical} and SART~\cite{andersen1984simultaneous}.
NeRF-based methods include IntraTomo~\cite{zang2021intratomo}, NAF~\cite{zha2022naf}, and SAX-NeRF~\cite{cai2024structure}, among which SAX-NeRF employs a Transformer-based architecture.
Gaussian-splatting methods include X-Gaussian~\cite{cai2025radiative}, which focuses on novel projection synthesis, and $\text{R}^2$-Gaussian~\cite{r2_gaussian}, which enables direct CT reconstruction via differentiable voxelization.

For dynamic CT reconstruction, we compare against recent methods designed for dynamic scenes, including HexPlane~\cite{cao2023hexplane}, which represents dynamic scenes via hexagonal decomposition, K-Planes~\cite{fridovich2023k}, an spatiotemporal radiance field representation, STNF4D~\cite{zhou2025spatiotemporal}, which  incorporates spatiotemporal information for dynamic CT, and 4D Gaussian Splatting (4DGS)~\cite{wu20244d}, which focuses on real-time rendering of dynamic scenes.

\begin{table}[t]
    \centering
    \setlength{\tabcolsep}{4pt}
    \renewcommand{\arraystretch}{1.05}
    \begin{tabular}{ccccc|cc}
        \toprule
        \multirow{2}{*}{Case} & \multicolumn{4}{c}{Component} & \multicolumn{2}{c}{Performance} \\
        \cmidrule(lr){2-5} \cmidrule(lr){6-7}
              & HE & STAB & MF & SR & PSNR $\uparrow$ & SSIM $\uparrow$ \\
        \midrule
        \multirow{2}{*}{Static} 
              & \cmark &        &        &        & 28.71 & 0.841 \\
              & \cmark &        &        & \cmark & 28.95 & 0.849 \\
        \midrule
        \multirow{4}{*}{Dynamic} 
              & \cmark &        &        &        & 34.26 & 0.939 \\
              & \cmark & \cmark &        &        & 34.89 & 0.945 \\
              & \cmark & \cmark & \cmark &        & 35.23 & 0.952 \\
              & \cmark & \cmark & \cmark & \cmark & \textbf{35.41} & \textbf{0.955} \\
        \bottomrule
    \end{tabular}
    \caption{Quantitative ablation of TG-Field components on static and dynamic situation.}
    \label{tab:component_ablation}
\end{table}

\begin{figure}[t]
    \centering
    \includegraphics[width=1\linewidth]{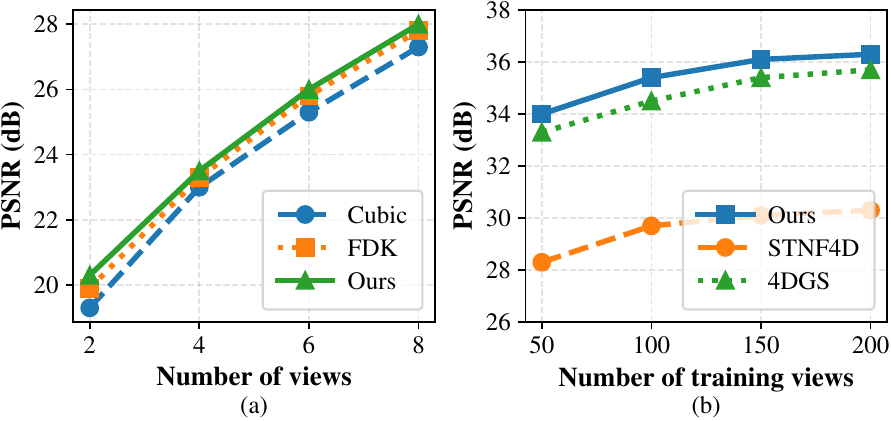}
    \caption{(a) Comparison of different initialization strategies under varying numbers of projections. (b) Reconstruction results of TG-Field using different numbers of projections.}
    \label{num_view}
\end{figure}

\subsection{Main Results}
\textbf{Static CT Reconstruction.}
Table~\ref{tab:ct_recon} reports quantitative comparisons for sparse-view CT reconstruction on synthetic and real datasets under 5, 10, and 20 views.
TG-Field achieves the best PSNR and SSIM in all settings.
On the synthetic dataset, it yields PSNR gains of about $0.5$, $0.8$, and $0.7$~dB over the strongest baseline for the 5-, 10-, and 20-view cases, respectively, together with consistent SSIM improvements.
On the real dataset, TG-Field similarly outperforms all baselines, with PSNR gains of about $0.7$, $0.9$, and $0.6$~dB for 5-, 10-, and 20-view settings.
Figure~\ref{view} compares the reconstruction quality of $\text{R}^2$-Gaussian and our TG-Field across different numbers of projection views.
Under extremely sparse-view conditions, $\text{R}^2$-Gaussian often yields blurred details and noticeable artifacts, whereas TG-Field preserves finer anatomical structures and effectively suppresses artifacts even with very limited projections.

\noindent\textbf{Dynamic CT Reconstruction.}
We evaluate TG-Field on three 4D CT datasets: XCAT, TCIA 4D-Lung, and SPARE.
As shown in Table~\ref{tab:dir_comparison}, TG-Field achieves the best PSNR/SSIM on all datasets and the highest average performance.
Compared with the strongest baseline 4DGS, it improves the average PSNR from $32.80$ to $33.78$~dB and SSIM from $0.933$ to $0.943$, indicating stronger robustness under severe view sparsity.
STNF4D performs reasonably but is more prone to noise and blurring, while HexPlane and K-Planes lag behind across all metrics.
4DGS is a strong baseline, yet TG-Field further enhances temporal modeling and local motion handling.
Visual results in Figure~\ref{4DCT} qualitatively confirm these gains, showing clearer structures and fewer artifacts.

\subsection{Ablation Study}

\textbf{Initialization Analysis.} To evaluate the impact of initialization quality on reconstruction performance, we compare our proposed geometry-informed strategy against two widely used alternatives: cubic uniform sampling and FDK-based point cloud initialization. Quantitative results under varying view numbers on the object CT are visualized in Figure~\ref{num_view}-(a). Our method consistently outperforms the baselines across 2, 4, 6, and 8-view settings, achieving notable gains in PSNR. These improvements confirm that accurate, geometry-aware initialization significantly enhances reconstruction fidelity under limited-view conditions.

\noindent\textbf{Projection Numbers.}
We further examine robustness with respect to the number of projections on the TCIA dataset.
As shown in Figure~\ref{num_view}-(b), reconstruction quality improves as more projections are available, and TG-Field consistently outperforms both STNF4D and 4DGS across all settings.

\noindent\textbf{Component Analysis.}
We perform ablations on both static and dynamic CT to quantify the contribution of each module in TG-Field, including the hash encoder (HE), spatiotemporal attention block (STAB), motion-flow module (MF), and semantic regularization (SR).
As summarized in Table~\ref{tab:component_ablation}, on static CT adding SR on top of HE brings a consistent gain in PSNR/SSIM, which is also reflected by sharper bone boundaries and fewer artifacts in Figure~\ref{Aba}.
On 4D CT, progressively introducing STAB and MF yields steady improvements, and the full model with SR achieves the best overall PSNR and SSIM.
Qualitative results in Figure~\ref{MF} further show that MF mainly refines motion-sensitive regions.

\section{Conclusion}
\label{conclusion}
We presented TG-Field, a geometry-aware Gaussian deformation framework for ultra-sparse-view tomographic reconstruction. It encodes geometric priors into radiative Gaussian optimization without requiring paired training data. Building on the same formulation, we handle dynamic CT via time-conditioned modeling to preserve temporal coherence. Experiments on synthetic and real datasets show consistent gains over recent methods, with higher fidelity and fewer artifacts in both static and dynamic settings.

\section{Acknowledgments}
We sincerely thank the anonymous reviewers for their constructive comments and Xueqi Ma for helpful discussions. This work was supported in parts by National Key R\&D Program of China (2024YFB3908500), NSFC (U21B2023. 52474105), Shenzhen Science and Technology Program (KJZD20240903100022028, KQTD20210811090044003, RCJC20200714114435012), Research Team Cultivation Program (2023QNT004) and Scientific Development Funds from Shenzhen University.

\bibliography{aaai2026}

\end{document}